\documentclass[conference, letterpaper, 10pt]{IEEEtran}
\IEEEoverridecommandlockouts
\usepackage{cite}
\usepackage[pdftex]{graphicx}
\usepackage{amsmath}
\usepackage{array}
\usepackage[caption=false,font=footnotesize]{subfig}
\usepackage{url}

\usepackage{pgfplots}
\usetikzlibrary{positioning}
\usepackage{color}

\usepackage{textcomp}
\usetikzlibrary{shapes,arrows}
\usetikzlibrary{automata,positioning} 

\usepackage{amssymb}
\usepackage{url}

\newcommand\Tstrut{\rule{0pt}{2.6ex}} 

\usepackage{tikz}
\usepackage{textcomp}
\usepackage{hyperref}
\usepackage{lipsum}

\newcommand\copyrighttext{%
  \footnotesize \textcopyright 2018 IEEE. Personal use of this material is permitted.
  Permission from IEEE must be obtained for all other uses, in any current or future
  media, including reprinting/republishing this material for advertising or promotional
  purposes, creating new collective works, for resale or redistribution to servers or
  lists, or reuse of any copyrighted component of this work in other works.
  DOI: \href{https://ieeexplore.ieee.org/document/8489656}{10.1109/IJCNN.2018.8489656}}
\newcommand\copyrightnotice{%
\begin{tikzpicture}[remember picture,overlay]
\node[anchor=south,yshift=10pt] at (current page.south) {\fbox{\parbox{\dimexpr\textwidth-\fboxsep-\fboxrule\relax}{\copyrighttext}}};
\end{tikzpicture}%
}

\begin{document}
%
\title{A Fully Attention-Based Information Retriever}



\author{\IEEEauthorblockN{Alvaro H. C. Correia, Jorge L. M. Silva, Thiago de C. Martins, Fabio G. Cozman}
\IEEEauthorblockA{Escola Politecnica - Universidade de Sao Paulo\\
Sao Paulo, Brazil
}
}
\maketitle
\copyrightnotice

\begin{abstract}
Recurrent neural networks are now the state-of-the-art in natural language processing because they can build rich contextual representations and process texts of arbitrary length. However, recent developments on attention mechanisms have equipped feedforward networks with similar capabilities, hence enabling faster computations due to the increase in the number of operations that can be parallelized. We explore this new type of architecture in the domain of question-answering and propose a novel approach that we call Fully Attention Based Information Retriever (FABIR). We show that FABIR achieves competitive results in the Stanford Question Answering Dataset (SQuAD) while having fewer parameters and being faster at both learning and inference than rival methods.
\end{abstract}


%
\IEEEpeerreviewmaketitle

\section{Introduction}
Question-answering (QA) systems that can answer queries expressed in natural language have been a perennial goal of the artificial intelligence community. An interesting strategy in the design of such systems is information extraction, where the answer is sought in a set of support documents. However, extracting information from large texts is still a challenging task, and most state-of-the-art models restrict themselves to single paragraphs. That is, in fact, the proposed focus of recent open-domain QA datasets, such as SQuAD \cite{Rajpurkar2016}. 

In SQuAD, each problem instance consists of a passage $\mathcal P$ and a question $\mathcal Q$. A QA system must then provide an answer $\mathcal A$ by selecting a snippet from $\mathcal P$. That format reduces the complexity of the task and also facilitates training, as one can learn a probability distribution over the words that compose the passage.
Since its publication in 2016, SQuAD has been targeted by many research groups, and the proposed models are gradually approaching (even overcoming) human-level performances.
All but a few of these models rely on Recurrent Neural Networks (RNNs), which currently dominate the state-of-the-art in most Natural Language Processing (NLP) tasks. However, RNNs do have some drawbacks, of which the most relevant to real-world applications is the high number of sequential operations, which increases the processing time of both learning and inference. 
To address these limitations, Vaswani et al. have proposed the Transformer, a machine translation model that introduces a new deep learning architecture solely based on ``attention'' mechanisms \cite{Vaswani}. We later clarify the meaning of attention in this context.

Inspired by the positive results of Vaswani et al. in machine translation, we have applied a similar architecture to the domain of question-answering, a model that we have named Fully Attention-Based Information Retriever (FABIR). Our goal then was to verify how much performance we can get exclusively from the attention mechanism, without combining it with several other techniques. We validated our model in the SQuAD dataset, which proved that FABIR not only achieves competitive results (F1:77.6\%, EM:67.7\%) but also has fewer parameters and is faster at both training and testing times than competing methods. Besides the development of a new architecture, we identify three major contributions of our work that have made these results possible:
\begin{itemize}
    \vspace{-\topsep}
    \item \textbf{Convolutional attention}: a novel attention mechanism that encodes many-to-many relationships between words, enabling richer contextual representations.
    \item \textbf{Reduction layer}: a new layer design that fits the pipeline proposed by Vaswani et al. \cite{Vaswani} and compresses the input embedding size for subsequent layers (this is especially beneficial when employing pre-trained embeddings).
    \item \textbf{Column-wise cross-attention}: we modify the cross-attention operation by \cite{Vaswani} and propose a new technique that is better suited to question-answering.
    \vspace{-\topsep}
\end{itemize}

This article is organized as follows. We first introduce some of the related work in question-answering and then present FABIR's architecture and its basic design choices. Subsequently, we report and comment our results in the SQuAD dataset. Finally, we compare the performance of FABIR with RNN-based models and draw some conclusions, suggesting directions for future work.

\section{Related Work}

The vast majority of papers that address the SQuAD dataset have adopted RNN-based models \cite{Hu2017,PanLZCCH17,Salant2017,Xiaodong2017,Yang2017,Hsin2017,  DBLP:journals/corr/abs-1711-00106,RiuLiu2017,bidaf+self-att,Zhequian2017,DBLP:journals/corr/GongB17,DBLP:journals/corr/ZhangZCDWJ17,DBLP:journals/corr/ShenHGC16,Chen2017,DBLP:journals/corr/LeeKP016,DBLP:journals/corr/WeissenbornWS17,DBLP:journals/corr/WangMHF16,DBLP:journals/corr/LiuHWYN17,Seo2016,Xiong2016,Wang16,DBLP:journals/corr/YangDYHCS16,DBLP:journals/corr/BahdanauBJGVB17,DBLP:journals/corr/YuZHYXZ16}. 
They all follow a similar pipeline, with pre-trained word-embeddings that are processed by bidirectional RNNs. Question and passage are processed independently, and their interaction is modeled by attention mechanisms \cite{Bahdanau2014} to produce an answer. There are slight differences in how each model employs attention, but they all calculate it over the hidden states of an RNN. Vaswani et al. were the first to apply attention directly over the word-embeddings, and thus derived a new neural network architecture which, without any recurrence, achieved state-of-the-art results in machine translation \cite{Vaswani}. In this section, we briefly discuss both types of attention models.

\subsection{Traditional Attention Mechanisms}
In recent years, attention mechanisms have been used with success in a variety of NLP tasks, such as machine translation \cite{Bahdanau2014,Vaswani} and natural language inference \cite{Rocktaschel16,WangJ15b}. Indeed, most models that target the SQuAD dataset use some form of attention to model the relationship between question and passage.

Attention can be defined as a mechanism that gives a score $\alpha_i$ to a vector $p_i$ from a set $P=[p_1,...,p_m]$ with respect to a vector $q_j$ from $Q=[q_1,...,q_n]$. This score is a function of both $P$ and $Q$ and is shown in its most general form in \eqref{eq:att}.

\begin{subequations}\label{eq:att}
\begin{equation}\label{eq:s}
s_{i,j} = f(p_i,q_j),
\end{equation}
\begin{equation}\label{eq:alpha_i}
\alpha_{i,j} = \frac{\exp(s_{i,j})}{\sum_{k=1}^{n}\exp(s_{i,k})},
\end{equation}
\end{subequations}
where $s_i$ and $\alpha_i$ are scalars and $f$ is a score function that measures the importance of $p_i$ relative to $q_{j}$.
Intuitively, a large weight $\alpha_i$ means that the vector $p_i$ is somehow strongly related to $Q$.
In the literature, two alternatives for $f$ have been proposed, additive \cite{Bahdanau2014} and multiplicative \cite{Luong2015} attentions:

\begin{equation}
f(p_i, q_{j}) =\left\{
\begin{aligned}
   W_3g(W_1 p_i + W_2 q_{j}) \quad \text{ (additive)}\\
   p_i^T W_1 q_{j} \quad \text{ (multiplicative)},
\end{aligned}
\right.
\label{eq:att_types}
\end{equation}
where $W_1$, $W_2$ and $W_3$ are learnable parameters and $g$ is a elementwise nonlinear function. For small vectors, additive and multiplicative attention mechanisms have been shown to produce similar results \cite{BritzGLL17}. 

In most models, the attention scores $\alpha$ are used to create a context vector $c$ given by a weighted sum of $P$, which is processed by an RNN:

\begin{subequations}\label{eq:RNN}
\begin{equation}
c^t = \sum_i \alpha_{i,t} p_i,
\end{equation}
\begin{equation}
 v^t = \mathrm{RNN}(v^{t-1}, c^t),
\end{equation}
\end{subequations}
where $v^t$ is the hidden state of the RNN at time $t$. Notably, in the SQuAD dataset, $P$ and $Q$ are the vectorial representations of passage and question, respectively.

\subsection{Google's Transformer}
The Transformer is a machine translation model introduced in \cite{Vaswani} that achieved state-of-the-art results by combining feedforward neural networks with a multiplicative attention mechanism applied over position-encoded embedding vectors. It defines three different matrices $U, K$ and $V$ that are associated with queries, keys, and values, respectively. Every attention operation in the Transformer is performed by multiplying these matrices as shown in \eqref{eq:att_parametrized}.

\begin{equation}\label{eq:att_parametrized}
\mathrm{att}(U,K,V) = \mathrm{softmax}\left(UW_{UK}K^T\right)V\;W_{V},
\end{equation}
where $W_{UK}, W_V \in \mathbb{R}^{d_{model}\times d_{model}}$ are weight matrices and $d_{model}$ is the embedding size of each word. 
Additionally, Vaswani et al. \cite{Vaswani} suggest a multi-head attention, in which $U,K$ and $V$ are divided into $n_{heads}$ heads and the attention in the $i^{th}$ head is computed as

\begin{subequations}\label{eq:multi_head}
\begin{equation}
    logits_i = UW_{U,i}(KW_{K,i})^T,
\end{equation}
\begin{equation}
    \mathrm{att_{i}}(U,K,V) = \mathrm{softmax}\left(logits_i\right)\,VW_{V,i},
\end{equation}
\end{subequations}
where  $W_{U,i}, W_{K,i}, W_{V,i} \in \mathbb{R}^{d_{model}\times d_{head}}$ are again learnable weight matrices and $d_{head}$ is the embedding dimension of each head. Finally, attention is computed by the concatenation of every head attention $att_i$, followed by an affine transformation:
\begin{equation}\label{eq:att_concat}
    \mathrm{att}(U,K,V)=\left[att_1;...;att_{n_{heads}}\right]W_O,
\end{equation}
where $W_{O}\in\mathbb{R}^{n_{heads}*d_{head}\times d_{model}}$.

If one wants to model the interdependence of words within a single piece of text, $U, K$ and $V$ are all equal and consist of the text of interest embedded in some vectorial space. This type of attention is often called {``}self-attention{''} or {``}self-alignment{''} \cite{Vaswani,Seo2016}. Conversely, if one seeks the relationship between words from two different passages, then $U$ represents one, while $K$ and $V$ represent the other. In that case, we talk about {``}cross-attention{''}.

\subsection{Other RNN-free Models}
We also identified another QA model \cite{anony-nonRNN} that is inspired by the architecture introduced by Vaswani et al. \cite{Vaswani}. Their model differs from ours in that it heavily relies on convolutions (46 layers against 2 in FABIR), which approximates it to other CNN NLP models \cite{GehringAGYD17}, rather than purely attention based models.
Although they report high F1 and EM scores (82.7\% and 73.3\%), our model is almost twice as fast in inference (259 samples/s against 440 in FABIR). Also, their model probably has a higher number of learned parameters due to the increased number of layers.

\section{FABIR}
In this section, we present FABIR's architecture and the main design decisions we have made to develop a lighter and faster question-answering model. In particular, we introduce the convolutional attention, the column-wise cross-attention, and the reduction layer, which build on the Transformer model \cite{Vaswani} to enable its application to question-answering.

\subsection{Embeddings}
We model each piece of text at the level of a word, i.e., sentences are defined by a sequence of vectors $\omega$, each one representing a word in a vectorial space $\mathbb{R}^{d_{input}}$. 
Thus, we build a new representation of question and passage to which we will refer as $\Omega_Q$ and $\Omega_P$, respectively. 

\begin{equation}
    \mathcal{P, Q} \xrightarrow{\text{embed}} \Omega_P \in \mathbb{R}^{P_{len} \times d_{input}}, \Omega_Q \in \mathbb{R}^{Q_{len} \times d_{input}},
\end{equation}
where $Q_{len}$ and $P_{len}$ denote the number of tokens in the question and the passage, respectively.
These embeddings are composed by word-level and character-level representations. The former is denoted by $\omega_{w}\in\mathbb{R}^{100}$ and was imported from the pre-trained embeddings of GloVe {``}6B{''} \cite{Pennington2014}. The latter is denoted by $\omega_c\in\mathbb{R}^{100}$ and is computed for each word as a result of the composition of its characters. 
Given a word with length $l$, $C=\left[c_1,c_2,...,c_l\right]$, in which $c_i\in\mathbb{R}^8$ are learned character embeddings, we compute $\omega_c$ by convolving $C$ with kernel $H \in \mathbb{R}^{ 1 \times 5 \times 8 \times 100}$ and applying max-over time pooling \cite{kim2016character}.
Finally, we squeeze $\omega_c$ values to $[-1,1]$ using a hyperbolic tangent activation function and pass the concatenation of $\omega_w$ and $\mathrm{tanh}(\omega_c)$ through a two-layer Highway-Network \cite{Srivastava2015} to obtain the final representation of a word.
\begin{equation}
    \omega=\mathrm{Highway}([\omega_w;\mathrm{tanh}\left(\omega_c\right)]).
\end{equation}

\subsection{Encoder}
In contrast to an RNN, FABIR does not process words in sequence, and hence needs to model the position of each word in a sentence differently. We add positional information to each word embedding using a trigonometric encoder as proposed in \cite{Vaswani}. Therefore, given a sequence of embedding vectors of even size $d_{model}$, the position of the $i^{th}$ word is encoded in a vector $e_i$ as follows:

\begin{equation}\label{eq:TrigEncoder}
e_i = \begin{bmatrix} \sin(i*f_1)\\\cos(i*f_1)\\...\\\sin(i*f_{d_{model}/2})\\\cos(i*f_{d_{model}/2})
\end{bmatrix},
\end{equation}
where $f_k$ are scalars, which were chosen according to \cite{Vaswani}.

The encoding of an embedding matrix $\Omega$ is represented by $E$ and the whole operation can be summarized as

\begin{equation}
    \mathcal{P, Q} \xrightarrow{\text{encode}} E_P \in \mathbb{R}^{P_{len} \times d_{model}}, E_Q \in \mathbb{R}^{Q_{len} \times d_{model}},
\end{equation}
where $d_{model}$ is the size of each position encoding, which is not necessarily equal to $d_{input}$.

The encoding $E$ can be summed to $\Omega$ to include the information of the position of each word in the text.
Indeed, in \cite{Vaswani}, the final vectorial representation of a piece of text is defined by the sum of the embeddings $\Omega$ with the position encoding $E$, which would require $d_{model}=d_{input}$.
However, we introduce a layer that processes embeddings and encodings separately before summing them up. Because we also use this layer to reduce the embedding size from $d_{input}$ to $d_{model}$, we named it {``}reduction layer{''}. The architecture of this type of layer is addressed further on.

\subsection{Convolutional Attention}
In FABIR the attention mechanism is inspired by the Transformer model introduced in \cite{Vaswani}. 
However, we hypothesize the word-to-word relationship in \eqref{eq:s} fails to capture the complexity of expressions involving groups of words.
To facilitate the modeling of the interdependence of surrounding words, we redefine $s_{i,j}$ as
\begin{equation}
s_{i,j} = f(p_{i-\frac{h-1}{2}}, ..., p_{i+\frac{h-1}{2}}, q_{j-\frac{w-1}{2}}, ..., q_{j+\frac{w-1}{2}}),
\end{equation}
where $h$ and $w$ are the height and width of a convolution kernel. This new type of attention, which we named {``}convolutional-attention{''}, is entirely defined by the following sequence of steps:
\begin{subequations}\label{eq:conv_attention}
\begin{equation}
    logits_i = UW_{U,i}(KW_{K,i})^T,
\end{equation}
\begin{equation} \label{eq:conv_padding}
    logits_{i, padded} = \mathrm{pad}(logits_i),
\end{equation}
\begin{equation}\label{eq:att_convolution}
    logits_{conv} = \mathrm{Conv}(logits_{padded}, H),
\end{equation}
\begin{equation}\label{eq:softmax_att}
    att_{head,i}= \mathrm{softmax}(logits_{conv,i})VW_{V,i},
\end{equation}
\begin{equation}
    \mathrm{att_{conv}}(U,K,V)= \left[att_{1};...; att_{n_{heads}}\right]W_O,
\end{equation}
\end{subequations}
where $\mathrm{Conv}$ represents a single convolutional layer with a trainable kernel $H\in \mathbb{R}^{h \times w \times n_{heads} \times n_{heads}}$ that has height $h$, width $w$, and number of filters and channels both equal to $n_{heads}$.
Note that in \eqref{eq:conv_padding} zero-padding is applied so that $logits_{conv}$ maintains the same dimension of $logits$.

\subsection{Sublayers}
\begin{figure*}[h!]
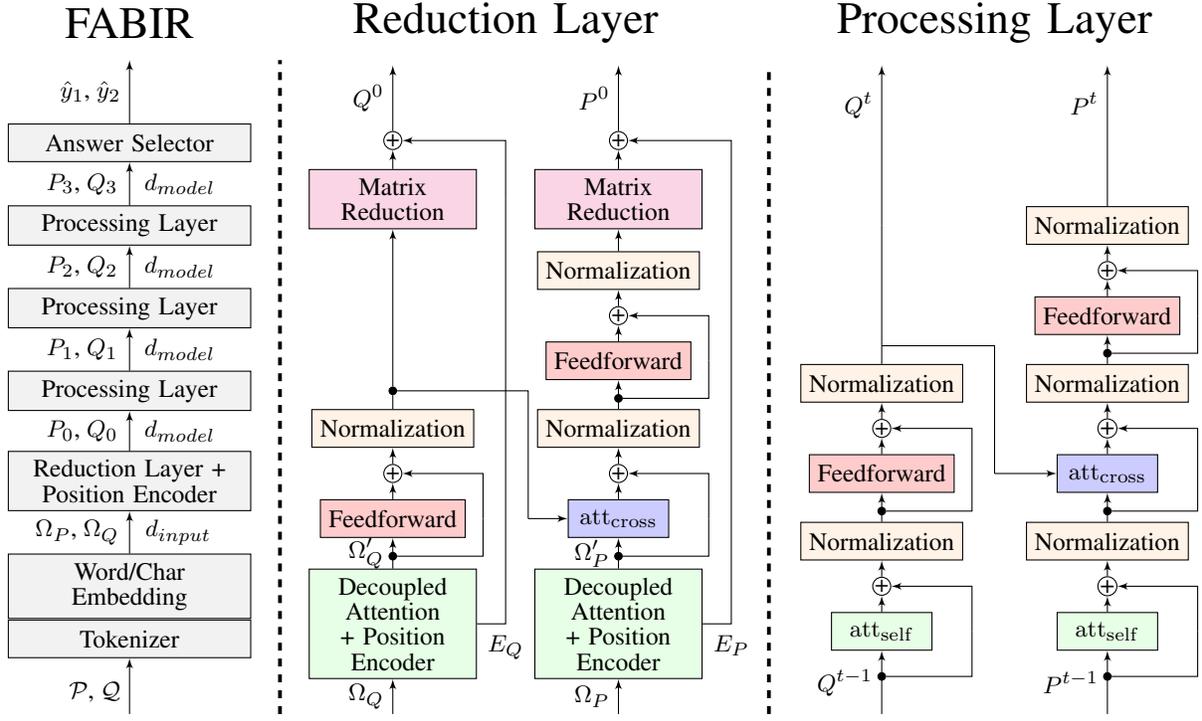

\centering
\include{layer_diagram}
\caption{On the left, a block diagram representation of FABIR, which receives as input the raw passage and question texts, $\mathcal{P}$ and $\mathcal{Q}$. The passage and question embedding matrices are $\Omega_P$ and $\Omega_Q$, respectively, and they both have an embedding dimension of $d_{input}$, which is a result of the tokenization, followed by the word/char embedding process. After the layer reduction, subsequent representations of $P$ and $Q$ have embedding size $d_{model}$ and already include the encoding of word positions. Finally, $\hat y_{1}$ and $\hat y_2$ are the indices, which define the answer to the passage-question pair $\mathcal{P}$ and $\mathcal{Q}$. On the center, a block representation of the reduction layer, which is the third layer in FABIR's pipeline. Finally, on the right side, it is the processing layer, which is similar to the reduction layer except by the absence of the {``}matrix reduction{''} sublayer and the substitution of the decoupled attention by a self-attention block.}\label{Fig:FabirDiagram}
\end{figure*}
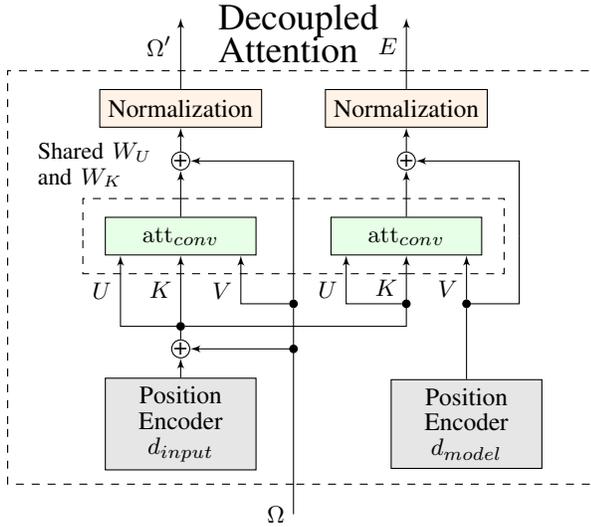
\begin{figure}[h!]
\centering
\tikzstyle{block} = [draw, fill=green!10, rectangle, 
    minimum height=1.5em, minimum width=6em]
\tikzstyle{block_encoder} = [draw, fill=black!10, rectangle, 
    minimum height=1.5em, minimum width=6em]
\tikzstyle{block_norm} = [draw, fill=orange!10, rectangle, 
    minimum height=1.5em, minimum width=4em]
\tikzstyle{block_FF} = [draw, fill=red!20, rectangle, 
    minimum height=1.5em, minimum width=4em]
\tikzstyle{block_cross} = [draw, fill=blue!20, rectangle, 
    minimum height=1.5em, minimum width=4em]
\tikzstyle{sum} = [draw, fill=black!0, circle, inner sep=0pt, node distance=1cm]
\tikzstyle{input} = [coordinate]
\tikzstyle{output} = [coordinate]
\tikzstyle{cross_line} = [draw, fill=black!100, circle, inner sep=0pt, minimum width=0.1cm, node distance=1cm]
\begin{tikzpicture}[auto, node distance=.7cm,>=latex']
 \fontsize{9.5pt}{9.5pt}\selectfont
    \node [coordinate, name=input] {U};
    \node [cross_line, above of=input, node distance = 2.2cm](split_part){};
    \node [sum, left of=split_part, node distance = 1.5cm](encoder-orig){\small{+}};
    \node [block_encoder, below of=encoder-orig, node distance = 1.0cm,text width=5em, align=center](encoder-block){Position Encoder $d_{input}$};
    \node [cross_line, above of=encoder-orig, node distance = 0.3cm](split_part2){};
    \node [block, above of=split_part2, node distance = 1.2cm](att_block){$\mathrm{att}_{conv}$};
    
    \node [sum, above of=att_block](sum_1){\small{+}};
    \node [block_norm, above of=sum_1] (norm1) {Normalization};
    \node [coordinate, above of=norm1, node distance = 1.2cm](output_emb){};
    \node [coordinate, right of=att_block, node distance = 1.5cm](ref3){};
    \node [cross_line, above of=split_part, node distance = 0.6cm](guide_V){};
    
    \node [block, right of=att_block, node distance=3.0cm](att_block2){$\mathrm{att}_{conv}$};
    
    \node [sum, above of=att_block2](sum_2){\small{+}};
    \node [block_norm, above of=sum_2] (norm2) {Normalization};
    \node [coordinate, above of=norm2, node distance = 1.2cm](output_enc){};
    \node [cross_line, below of=att_block2, node distance=0.9cm](ref1){};
    \node [block_encoder, right of=encoder-block, node distance=3.8cm, text width=5em, align=center] (encoder-red) {Position Encoder $d_{model}$};
    \node [cross_line, above of=encoder-red, node distance=1.6cm](ref2){};
    \node [coordinate, right of=att_block2, node distance = 1.5cm](ref4){};

    \draw [->] (input) |- node [pos=0]{$\Omega$}(encoder-orig);
    \draw [->] (ref3) |- (sum_1);
        \draw [->] (sum_1) -- (norm1);
        \draw [->] (norm1) -- node[pos=0.6]{$\Omega'$} (output_emb);
    \draw [->] (att_block) -- (sum_1);
\draw[->]  (guide_V)-| node [pos=0.64]{$V$} ([xshift=.8cm]att_block.south);
\draw [-] (split_part) -- (guide_V);
\draw [-] (encoder-orig) -- (split_part2);
\draw [-] (guide_V) -- (ref3);
\draw[->]  (split_part2)-| node [pos=0.75]{$U$} ([xshift=-.8cm]att_block.south);
\draw[->]  (split_part2)-| node [pos=0.74]{$K$} ([xshift=0cm]att_block.south);
\draw[->] (encoder-block)-- (encoder-orig);

\draw[->] (ref1)-| node [pos=0.65]{$U$} ([xshift=-.8cm]att_block2.south);
\draw[->] (split_part2)-| node [pos=0.77]{$K$} ([xshift=0cm]att_block2.south);
\draw [-] (encoder-red) -- (ref2);
\draw[->] (ref2)-| node [pos=0.62]{$V$} ([xshift=+.8cm]att_block2.south);
\draw[-] (ref2)-| (ref4);
\draw[->] (att_block2) -- (sum_2);
\draw[->] (ref4) |- (sum_2);
\draw [->] (sum_2) -- (norm2);
\draw [->] (norm2) -- node[pos=0.6]{$E$} (output_enc);
\node [coordinate, above of=att_block, node distance=0.95cm](msg_up){};
 \node [left of=msg_up, node distance = 0.9cm, text width=2cm](txt) {\small{Shared $W_U$ and $W_K$}};
    \node [coordinate,left of=att_block, node distance=1.3cm](p1){};
    \node [coordinate,above of=att_block, node distance=.5cm](p2){};
    \node [coordinate,right of=att_block2, node distance=1.3cm](p3){};
    \node [coordinate,below of=att_block2, node distance=.5cm](p4){};
    \draw [-, dashed] (p1) |- (p2);
    \draw [-, dashed] (p2) -| (p3);
    \draw [-, dashed] (p3) |- (p4);
    \draw [-, dashed] (p4) -| (p1);
    
  \node [right of=norm1, node distance = 2.0cm](txt_1){};
 \node [above of=txt_1, node distance = 1.0cm, text width=3cm](txt) {\Large{Decoupled Attention}}; 
    \node [coordinate,above of=norm2, node distance=.5cm](p5){};
    \node [coordinate,right of=norm2, node distance=2.5cm](p6){};
    \node [coordinate,above of=input, node distance=0.4cm](p7){};
    \node [coordinate,left of=encoder-block, node distance=2.3cm](p8){};
    \draw [-, dashed] (p5) -| (p6);
    \draw [-, dashed] (p6) |- (p7);
    \draw [-, dashed] (p7) -| (p8);
    \draw [-, dashed] (p8) |- (p5);

\end{tikzpicture}
\caption{Decoupled Attention. In contrast to previous attention mechanisms,  this structure computes the embedding matrix $\Omega'\in\mathbb{R}^{\Omega_{len} \times d_{input}}$ and the encoder matrix $E\in\mathbb{R}^{\Omega_{len} \times d_{model}}$ separately. Shared weights $W_U$ and $W_K$ and same inputs $U$ and $K$ allow us to compute the convolutional attention pipeline only once until \eqref{eq:att_convolution}, as in self-attention. }\label{Fig:decoupled_self_att}
\end{figure}
After converting question and passage to their vectorial representation $Q$ and $P$, we apply a series of operations that we call sublayers. In this section, we introduce each of these operations.

\subsubsection{Self-attention}
Self-attention ($\mathrm{att_{self}}$) is the mechanism that models the interdependence between words in the same piece of text. 
It has been proven to help relating distant words, which is crucial to understand the long sentences that appear in context paragraphs in SQuAD.
In FABIR, self-attention is a sublayer that applies such operation via convolutional attention and is defined as
\begin{equation}\label{eq:self_att_def}
\mathrm{att_{self}}(P) = \mathrm{att_{conv}}(P,P,P).
\end{equation}
\subsubsection{Column-wise Cross-attention}
Cross-attention ($\mathrm{att_{cross}}$) differs from other types of attention by relating two different pieces of text.
Given $P$ and $Q$, cross-attention of $Q$ over $P$ is defined as
\begin{equation}\label{eq:cross_att_def}
\mathrm{att_{cross}}(P,Q) = \mathrm{att_{conv}}(P,Q,Q).
\end{equation}

In contrast to Vaswani et al. \cite{Vaswani}, where  the softmax in \eqref{eq:softmax_att} is applied in a row-wise manner, we suggest column-wise cross-attention. More precisely, we sum over $i$ instead of $j$ in \eqref{eq:alpha_i}. 
Row-wise softmax is inadequate in QA because, in practice, it computes a weighted average of the question words for every passage word, and thus cannot model the likely scenario in which not every word in the passage is related to the question. In contrast, the column-wise softmax attributes greater weights to passage words that are more closely related to the respective question word, which seems appropriate for the SQuAD task.

Many question-answering models employ cross-attention in both directions: $att_{cross}(P,Q)$ and $att_{cross}(Q,P)$ \cite{Yang2017,Xiong2016,Seo2016}. However, in FABIR we have observed better results when only the former is used.

\subsubsection{Feedforward}
The feedforward sublayer is solely composed of a neural network with a single hidden layer, which is applied vector-wise.
Following the architecture suggested by Vaswani et al. \cite{Vaswani}, the feedforward sublayer is implemented in \eqref{eq:FF} with a two-layer neural network:
\begin{equation}\label{eq:FF}
    x_{i,out} = \mathrm{ReLU}\left(x_i\;W_{1} +b_{1}\right)W_{2}+b_{2},
\end{equation}
where $W_{1} \in \mathbb{R}^{d_{model}\times d_{hidden}}$, $W_{2} \in \mathbb{R}^{d_{hidden}\times d_{model}}$, $b_1 \in \mathbb{R}^{1\times  d_{hidden}}$ and $b_2\in \mathbb{R}^{1\times d_{model}}$ are all trainable parameters, $d_{hidden}$ is the dimension of the hidden layer in \eqref{eq:FF} and $\mathrm{ReLU}(x) = \max(0, x)$.

\subsubsection{Normalization}
This operation is also applied vector-wise and it normalizes the embedding of each word so that its variance and mean are reduced to 1 and 0, respectively.
The primary goal of layer normalization is to accelerate training as shown in \cite{IoffeS15,Hinton16}. 

\subsection{Layers}
A layer $\mathcal{L}$ is a combination of sublayers that produce a transformation in the representation of question and passage:

\begin{equation}
    P^{l+1}, Q^{l+1}=\mathcal{L}(P^l, Q^l).
\end{equation}

Typically, a layer is composed of self-attention with shared weights applied to $P$ and $Q$ individually, followed by cross-attention and feedforward sublayers. 
This standard layer is called {``}processing layer{''} and is illustrated in Figure \ref{Fig:FabirDiagram}.
Note that, to facilitate training, every sublayer is followed by normalization.

FABIR is formed by stacking layers on top of each other as shown in Figure \ref{Fig:FabirDiagram}. Not counting the pre-processing and answer selector layers, our best performing model was composed of four layers, of which the last three are processing layers and the first is a reduction layer.

\subsection{Reduction Layer}
The SQuAD dataset is relatively small for the training of word embeddings, and pre-trained word vectors have been favored in the literature \cite{Seo2016}. Nonetheless, we observed that the new architecture introduced by Vaswani et al. \cite{Vaswani} is more susceptible to overfitting than RNNs when presented with large embedding sizes. Hence, we needed a method to compress the word representations, and thus facilitate and speed up training by reducing the number of parameters.
A straightforward method to reduce the input embedding size is to multiply it by a matrix with the required dimensions:
\begin{equation}\label{eq:matrix_reduction}
    \omega_{model}=W_{Reduction}\omega_{input},
\end{equation}
where $\omega_{model}\in\mathbb{R}^{1\times d_{model}}$, $\omega_{input}\in\mathbb{R}^{1\times d_{input}}$ are the embedding vectors, and $W_{Reduction}\in\mathbb{R}^{d_{model}\times d_{input}}$ is a weight matrix, to which we refer as {``}reduction matrix{''}. 

Although matrix reduction is quite simple, it discards information before any processing and hence might hamper performance by preventing the network from using some relevant data.
To incorporate that information before discarding it, we could add a large processing layer followed by a matrix reduction, but our experiments have shown that this approach does not yield positive results in FABIR.
Our interpretation of that behavior is that the position encoding is somehow dissolved in the matrix reduction process. 

A possible solution is to process embeddings of size $d_{input}$ and encodings of size $d_{model}$ independently and thus limit the reduction operation to the embeddings. We then suggest decoupled attention, a mechanism that allows us to apply self-attention to embeddings and encodings separately to preserve their different sizes, as described in Figure \ref{Fig:decoupled_self_att}.
Both operations use the full embedding $\Omega\in\mathbb{R}^{\Omega_{len}\times d_{input}}$, but the decoupled attention outputs embeddings with size $d_{input}$ and encodings with final size $d_{model}$. 

After applying decoupled attention with shared weights in $P$ and $Q$, we add a full processing layer for the embeddings $\Omega'$ with size $d_{input}$. That layer is equivalent to a regular processing layer $\mathcal{L}$, but it processes only $\Omega'$, leaving the encoding $E$ untouched.
Finally, we use a reduction matrix to scale $\Omega'$ down to $d_{model}$ and add the encoder matrix $E$ of same size, which is the output from the decoupled attention sublayer.
Figure \ref{Fig:FabirDiagram} describes this whole process, which we named {``}reduction layer{''}.

\subsection{Answer Selection}
Given that in SQuAD the answer is always contained in the supporting paragraph $\mathcal{P}$, the output of the model is merely the indices $\hat{y}_1$ and $\hat{y}_2$ that represent the first and the last word of the answer, respectively. Therefore, we can model the answer as two probability distributions $\pi_1$ and $\pi_2$ over the passage $\mathcal{P}$, and train the model to minimize the negative log-likelihood. Given the true indices $y_1$ and $y_2$, the cost function is then defined as follows:

\begin{equation}\label{eq:negativelikelihood}
    J = -(y_1 \log(\pi_1)+ y_2 \log(\pi_2)).
\end{equation}

To compute the cost function, we apply a two layered convolutional neural network with hidden layer size 32 and output size 2, as we need one dimension for each probability distribution. Both convolutions have kernel size 9 and the activation function (ReLU) is applied only after the first layer. Subsequently, each output dimension  passes through a softmax operation to compute the probability distributions $\hat\pi^{y_1}$ and $\hat\pi^{y_2}$. Finally, selecting the indices $\hat y_{1}$ and $\hat y_2$ becomes an optimization problem:

\begin{equation}\label{eq:opt_index}
\begin{aligned}
& \underset{i,j}{\text{maximize}}
& & \hat\pi^{y_1}_i*\hat\pi^{y_2}_j \\
& \text{subject to}
& & i \leq j < i+15,
\end{aligned}
\end{equation}
where $15$ represents the maximum allowed  answer length. This superior limit is imposed to avoid long answers, since short answers are more frequent.

\section{Experimental Results}
We have trained our FABIR model during 54 epochs with a batch size of 75 in a GPU NVidia Titan X with 12 GB of RAM. We developed our model in Tensorflow \cite{GoogleResearch2015} and made it available at 
\url{https://worksheets.codalab.org/worksheets/0xee647ea284674396831ecb5aae9ca297/} for replicability. 

We pre-processed the texts with the NLTK Tokenizer \cite{Bird2009}. As suggested in \cite{Vaswani}, we have chosen the Adam optimizer \cite{Kingma2014a} with the same hyperparameters, except for the learning rate, which was divided by two in our implementation.  For regularization, we applied residual and attention dropout \cite{Vaswani} of 0.9 in processing layers and of 0.8 in the reduction layer. In the character-level embedding process, a dropout of 0.75 was added before the convolution. Additionally, a dropout of 0.8 was added before each convolutional layer in the answer selector. We set processing layers dimension $d_{model}$ to 100, the number of heads $n_{heads}$ in each attention sublayer to 4, the feed-forward hidden size to 200 in processing layers and 400 in the reduction layer. Convolution kernels in attention sublayers had spatial dimensions $1\times 5$.

\subsection{Architecture Evaluation}
To better evaluate FABIR's architecture, we ran controlled tests on each of its key elements. Table \ref{tab:tests} show the results of these experiments regarding the F1 and EM scores, and the Training Time (TT) over 18-epoch runs. This analysis confirms the effectiveness of char-embeddings, as its addition increased the F1 and EM scores, by 2.7\% and 3.1\%, respectively. Most importantly, when the convolutional attention was replaced by the standard attention mechanism proposed in \cite{Vaswani}, the performance dropped by 2.4\% in F1 and 2.5\% in EM. That validates the contribution of this new attention method in building elaborate contextual representations. Moreover, the tests also indicate that the reduction layer is capable of producing useful word representations when compressing the embeddings. Indeed, when we replaced that layer by a standard feedforward layer with the same reduction ratio, there was a drop of 2.1\% and 2.5\% in the F1 and EM scores, respectively. Finally, we observed that the column-wise cross-attention outperforms its row-wise counterpart by 2.0\% and 1.9\% in F1 and EM, respectively. It confirms the intuition that applying the softmax over the passage words is more adequate in QA.

The training times indicate that each one of the new mechanisms introduced in FABIR incurred an increase in the processing cost. Nonetheless, that was outweighed by the improvement in performance, especially because FABIR is still significantly faster than competing RNN models.

\begin{table}[h!]
    \centering
    \caption{Architecture variations. F1 and EM scores in the dev set.}
    \begin{tabular}{| p{4.5cm} ||p{0.8cm}|p{0.8cm}|p{0.7cm}|}
        \hline\Tstrut
        \textbf{Architecture} &\textbf{F1(\%)} &\textbf{EM(\%)} & \textbf{TT} \\ [0.5ex] 
        \hline
        \hline
         FABIR & 75.6 &  65.1& 2h14m
        \\\hline 
        FABIR without char embedding & 72.9 & 62.0  & 1h48m
        \\\hline FABIR without convolutional attention & 73.2 & 62.6  & 1h49m
        \\\hline FABIR without the reduction layer & 73.5& 62.6 & 1h59m
        \\\hline FABIR with row-wise cross attention & 73.6 & 63.2 & 2h08m
        \\\hline FABIR with 2 attention heads & 73.8 & 63.2  & 2h12m
        \\\hline FABIR with linear answer selector & 74.0 & 62.8 & 2h10m
        \\\hline FABIR with 4 layers $\mathcal{L}_{Q\rightarrow X}$ & 75.5 & 64.7  & 2h47m
        \\\hline FABIR with 2 layers $\mathcal{L}_{Q\rightarrow X}$ & 75.0 & 64.1  & 1h55m
        \\\hline
    \end{tabular}
    \label{tab:tests}
\end{table}

\subsection{FABIR vs BiDAF}
This section compares our model to traditional RNN-based question-answering models.
To have a comprehensive comparison, we took a state-of-the-art model \cite{Seo2016} developed in Tensorflow \cite{GoogleResearch2015} that had its code openly available at \url{https://github.com/allenai/bi-att-flow}. That way, we could run our experiments with both models in the same piece of hardware to have a fair comparison between them. In Table \ref{tab:BidafvsFabir}, the BiDAF scores without parentheses were achieved after training their model for 18,000 iterations of batch size 60 in our hardware. Conversely, values in parentheses are BiDAF's official scores in the SQuAD ranking \cite{squad-website}.
Note that for both models, we batch the examples by paragraph length to improve computational efficiency.

\begin{table}[h!]
    \centering
    \caption{Comparison between FABIR and BiDAF \cite{Seo2016} models. }
    \begin{tabular}{| c || c |c|}
        \hline\Tstrut
         & \textbf{FABIR} &\textbf{BiDAF} \\ [0.5ex] 
        \hline
        \hline\Tstrut
         \# of Training Variables & \textbf{1,385,198} & 2,695,851\\
         Inference Sample/Second & \textbf{440} & 78 \\
         Training Sample/Second &\textbf{202} & 45 \\
         Training Time/Epoch &\textbf{7m13s} & 32m30s \\
         Training Epochs & 54  & \textbf{12} \\
         Training Time & 6h30m  & 6h30m \\
         F1 in the dev set (\%) & \textbf{77.6} & 77.0 (77.3)\\
         EM in the dev set (\%) & 67.6 &67.3 (\textbf{68.0})\\\hline 
    \end{tabular}
    \label{tab:BidafvsFabir}
\end{table}

Regarding EM and F1 scores, FABIR and BiDAF showed similar performances. 
Their similar scores render further comparisons even more telling, because their differences cannot be explained by their overall performances, but exclusively by their architectures.
Although both models required similar training times to reach these scores, the time for training one epoch in FABIR was more than four times shorter, which could be useful for tackling larger data sets. 

Concerning inference time, FABIR was more than five times faster in processing the 10,570 question-passage pairs in the development data set. 
FABIR's faster inference is a substantial advantage in large-scale applications, such as information extraction in large corpora.
Indeed, when running applications such as search tools or user interfaces, the inference time is critical to tackle real-world problems. 
Concerning the number of training variables, FABIR has almost 50\% fewer parameters than BiDAF, which incurs two major advantages. First, its training time is expected to be shorter, because the number of variables to be updated in every iteration is smaller. Secondly, it has lower memory requirements, which is attractive to applications that dispose of low computational power.
\subsection{FABIR and BiDAF Statistics}
In this section we analyze the performance of FABIR and BiDAF in the different types of question in SQuAD.

        \begin{figure}[b!]
            \centering
\begin{tikzpicture}[scale=.9]
\begin{axis}[
width=9.5cm,
height=4.0cm,
xlabel={Answer Length},
ylabel={F1-Score (\%)},
xmin=-0.5, xmax=10.5,
ymin=55, ymax=90,
xtick={0,1,2,3,4,5,6,7,8,9,10},
xticklabels={1,,3,,5,,7,,9,,10+},
tick align=outside,
xticklabel style = {rotate=0},
tick pos=left,
xmajorgrids,
x grid style={white!69.019607843137251!black},
ymajorgrids,
y grid style={white!69.019607843137251!black},
legend cell align={left},
legend style={draw=white!80.0!black},
legend entries={{FABIR},{BiDAF}}
]
\addlegendimage{mark=diamond*, violet}
\addlegendimage{mark=*, orange}
\addplot [semithick, violet, mark=diamond*, mark size=3, mark options={solid}, only marks]
table {%
0 77.8659005801275
1 79.2681073131318
2 78.183012972056
3 77.8758941862664
4 76.1137114319913
5 74.4846240986424
6 71.9065592024469
7 69.2167689407319
8 70.7281482130149
9 68.4994093690848
10 60.31290123731
};
\addplot [semithick, violet]
table {%
0 77.8659005801275
1 79.2681073131318
2 78.183012972056
3 77.8758941862664
4 76.1137114319913
5 74.4846240986424
6 71.9065592024469
7 69.2167689407319
8 70.7281482130149
9 68.4994093690848
10 60.31290123731
};
\addplot [semithick, orange, mark=*, mark size=2, mark options={solid}, only marks]
table {%
0 77.3226070566798
1 78.8947394946639
2 77.2187234626416
3 75.8347888386595
4 74.89354458724
5 72.5785613756593
6 69.8951604652573
7 70.2865440977904
8 69.7171868928852
9 72.144785314245
10 67.3925496525886
};
\addplot [semithick, orange]
table {%
0 77.3226070566798
1 78.8947394946639
2 77.2187234626416
3 75.8347888386595
4 74.89354458724
5 72.5785613756593
6 69.8951604652573
7 70.2865440977904
8 69.7171868928852
9 72.144785314245
10 67.3925496525886
};

\end{axis}

\end{tikzpicture}
            \caption{F1-Score against answer length for FABIR and BiDAF.}
            \label{fig:ans_len_stat}
        \end{figure}
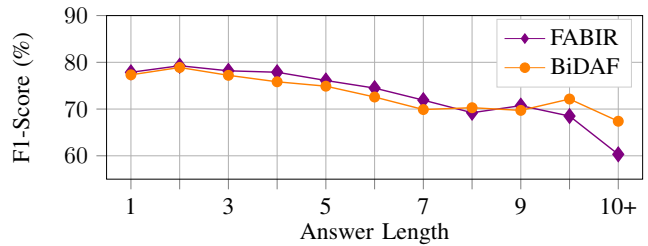
        \begin{figure}[b!]
            \centering
\begin{tikzpicture}[scale=.9]

\begin{axis}[
xlabel={Question Length},
width=9.5cm,
height=4.0cm,
xmin=-0.2, xmax=4.2,
ymin=55, ymax=90,
xtick={0,1,2,3,4},
ylabel={F1-Score (\%)},
xticklabels={7,11,15,19,19+},
xticklabel style = {rotate=0},
tick align=outside,
tick pos=left,
xmajorgrids,
x grid style={white!69.019607843137251!black},
ymajorgrids,
y grid style={white!69.019607843137251!black},
legend cell align={left},
legend style={draw=white!80.0!black},
legend entries={{FABIR},{BiDAF}},
legend pos=south east
]
\addlegendimage{mark=diamond*, violet}
\addlegendimage{mark=*, orange}
\addplot [semithick, violet, mark=diamond*, mark size=3, mark options={solid}, only marks]
table {%
0 78.2214313598784
1 77.8170469833055
2 77.6536057514549
3 76.1360947639036
4 76.9344496499669
};
\addplot [semithick, violet]
table {%
0 78.2214313598784
1 77.8170469833055
2 77.6536057514549
3 76.1360947639036
4 76.9344496499669
};
\addplot [semithick, orange, mark=*, mark size=2, mark options={solid}, only marks]
table {%
0 76.3724855804995
1 76.6935700696069
2 77.3841048117413
3 77.5611838954533
4 78.433931811787
};
\addplot [semithick, orange]
table {%
0 76.3724855804995
1 76.6935700696069
2 77.3841048117413
3 77.5611838954533
4 78.433931811787
};

\end{axis}

\end{tikzpicture}
            \caption{F1-Score against question length for FABIR and BiDAF.}
            \label{fig:q_len_stat}
        \end{figure}
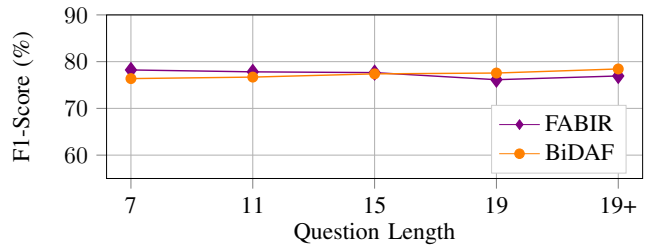
Figure \ref{fig:ans_len_stat} shows that shorter answers are easier for both models: while they reach more than 75\% F1 for answers that are shorter than four words, for answers longer than ten words these scores drop to 60.4\% and 67.3\% for FABIR and BiDAF, respectively. 
Long answers are not only more challenging but are also underrepresented in the dataset, which introduces a bias towards short responses. More than 79\% of the answers in SQuAD have five words or fewer.

In contrast to what has been observed for answers, longer questions seem not to increase the complexity of the task. 
In Figure \ref{fig:q_len_stat}, the F1 scores for both models varied by less than 2.5\% in the considered question length intervals.

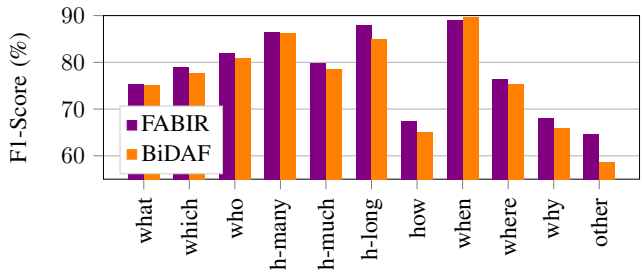
\begin{figure}[t!]
            \centering
\begin{tikzpicture}[scale=.9]

\begin{axis}[
xmin=-0.71, xmax=11.06,
ymin=55, ymax=90,
width=9.5cm,
height=4.0cm,
ylabel={F1-Score (\%)},
xtick={0.175,1.175,2.175,3.175,4.175,5.175,6.175,7.175,8.175,9.175,10.175},
xticklabels={what,which,who,h-many,h-much,h-long,how,when,where,why,other},
tick align=outside,
xticklabel style = {rotate=90},
tick pos=left,
x grid style={white!69.019607843137251!black},
y grid style={white!69.019607843137251!black},
ymajorgrids,
legend style={draw=white!80.0!black},
legend cell align={left},
legend pos=south west,
legend entries={{FABIR},{BiDAF}}
]
\addlegendimage{only marks, mark=square*, violet}
\addlegendimage{only marks, mark=square*, orange}
\draw[fill=violet,draw opacity=0] (axis cs:-0.175,0) rectangle (axis cs:0.175,75.4161350587143);
\draw[fill=violet,draw opacity=0] (axis cs:0.825,0) rectangle (axis cs:1.175,78.877619335507);
\draw[fill=violet,draw opacity=0] (axis cs:1.825,0) rectangle (axis cs:2.175,81.9361505866093);
\draw[fill=violet,draw opacity=0] (axis cs:2.825,0) rectangle (axis cs:3.175,86.4603350146482);
\draw[fill=violet,draw opacity=0] (axis cs:3.825,0) rectangle (axis cs:4.175,79.7459661550571);
\draw[fill=violet,draw opacity=0] (axis cs:4.825,0) rectangle (axis cs:5.175,87.9159302236225);
\draw[fill=violet,draw opacity=0] (axis cs:5.825,0) rectangle (axis cs:6.175,67.3460362673039);
\draw[fill=violet,draw opacity=0] (axis cs:6.825,0) rectangle (axis cs:7.175,88.9908242669437);
\draw[fill=violet,draw opacity=0] (axis cs:7.825,0) rectangle (axis cs:8.175,76.3114379227722);
\draw[fill=violet,draw opacity=0] (axis cs:8.825,0) rectangle (axis cs:9.175,68.0391203927265);
\draw[fill=violet,draw opacity=0] (axis cs:9.825,0) rectangle (axis cs:10.175,64.5917883286304);
\draw[fill=orange,draw opacity=0] (axis cs:0.175,0) rectangle (axis cs:0.525,75.0649437981982);
\draw[fill=orange,draw opacity=0] (axis cs:1.175,0) rectangle (axis cs:1.525,77.7551208492504);
\draw[fill=orange,draw opacity=0] (axis cs:2.175,0) rectangle (axis cs:2.525,80.8563146155239);
\draw[fill=orange,draw opacity=0] (axis cs:3.175,0) rectangle (axis cs:3.525,86.246119517685);
\draw[fill=orange,draw opacity=0] (axis cs:4.175,0) rectangle (axis cs:4.525,78.4784299394689);
\draw[fill=orange,draw opacity=0] (axis cs:5.175,0) rectangle (axis cs:5.525,84.897918995085);
\draw[fill=orange,draw opacity=0] (axis cs:6.175,0) rectangle (axis cs:6.525,64.9525557546168);
\draw[fill=orange,draw opacity=0] (axis cs:7.175,0) rectangle (axis cs:7.525,89.7150384949179);
\draw[fill=orange,draw opacity=0] (axis cs:8.175,0) rectangle (axis cs:8.525,75.2185234271493);
\draw[fill=orange,draw opacity=0] (axis cs:9.175,0) rectangle (axis cs:9.525,65.8488894220867);
\draw[fill=orange,draw opacity=0] (axis cs:10.175,0) rectangle (axis cs:10.525,58.7057009210132);
\end{axis}

\end{tikzpicture}
            \caption{F1-Score against question type for FABIR and BiDAF.}
            \label{fig:q_type_stat}
            
        \end{figure}

Question type is a strong predictor of performance. 
Figure \ref{fig:q_type_stat} shows that both models had their best performance with {``}when{''} questions. Answers to these types of questions are easier to infer because they are usually composed of time-related words, such as months, years, seasons or weekdays, which are easier to distinguish from the rest of the text.
Together with {``}when{''} questions, {``}how long{''} and {``}how many{''} also proved easier to respond, as they possess the same property of having a smaller universe of possible answers. 
In contrast to these, {``}how{''} and {``}why{''} questions resulted in considerably lower F1 and EM scores, as they can be answered by any sentence, and hence require a deeper understanding of the text. 
Note that {``}how{''} questions do not include {``}how many{''}, {``}how much{''} or {``}how long{''} questions, whose answers are more predictable. 

{``}Other{''} questions include alternatives, such as {``}Name a type of...{''} or {``}Does it...{''} or even questions with typos, such as {``}Hoe was ...{''}, which should be {``}How was ...{''}. These questions are more challenging because they might have multiple correct answers and require higher levels of abstraction.
For instance, to respond to a question such as {``}Name an ingredient...{''}, the model would need a deep understanding of the semantics of the word {``}ingredients{''} to identify  {``}tomatoes{''} or {``}cheese{''} as possible answers.
Questions which expect a {``}yes{''} or a {``}no{''} as an answer are also difficult because it is not always possible to find those words in a snippet from the passage.

        \begin{figure}[t!]
            \centering
\begin{tikzpicture}[scale=.9]

\begin{axis}[
xlabel={Passage Length},
width=9.5cm,
height=4.0cm,
ylabel={F1-Score (\%)},
xmin=-0.3, xmax=6.3,
ymin=55, ymax=90,
xtick={0,1,2,3,4,5,6},
xticklabels={60,100,140,180,220,260,260+},
tick align=outside,
xticklabel style = {rotate=0},
tick pos=left,
xmajorgrids,
x grid style={white!69.019607843137251!black},
ymajorgrids,
y grid style={white!69.019607843137251!black},
legend cell align={left},
legend style={draw=white!80.0!black},
legend entries={{FABIR},{BiDAF}},
legend pos=south east
]
\addlegendimage{mark=diamond*, violet}
\addlegendimage{mark=*, orange}
\addplot [semithick, violet, mark=diamond*, mark size=3, mark options={solid}, only marks]
table {%
0 73.8806962898475
1 77.4428638839503
2 77.6337618535844
3 77.8934405177054
4 78.2595008779401
5 79.4213473163261
6 75.9936677862175
};
\addplot [semithick, violet]
table {%
0 73.8806962898475
1 77.4428638839503
2 77.6337618535844
3 77.8934405177054
4 78.2595008779401
5 79.4213473163261
6 75.9936677862175
};
\addplot [semithick, orange, mark=*, mark size=2, mark options={solid}, only marks]
table {%
0 70.8261035064209
1 77.7064335633814
2 76.662242838158
3 77.4687840919546
4 78.0171021587814
5 78.5544568796706
6 75.9194274982013
};
\addplot [semithick, orange]
table {%
0 70.8261035064209
1 77.7064335633814
2 76.662242838158
3 77.4687840919546
4 78.0171021587814
5 78.5544568796706
6 75.9194274982013
};
\end{axis}

\end{tikzpicture}
            \caption{F1-Score against passage length for FABIR and BiDAF.}
            \label{fig:par_len_stat}
        \end{figure}
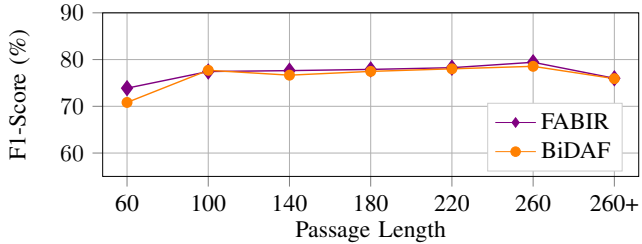

Figure \ref{fig:par_len_stat} shows the performance of FABIR and BiDAF against the passage length. It is curious that shorter passages showed the worst performance for both models. It is hard to interpret that result as, intuitively, one would expect brief passages to be easier to interpret. One possible explanation is that short passages give fewer options of simple questions, such as {``}when{''}, {``}who{''}, or {``}how many{''}, and the annotators of the dataset had to resort to more elaborate alternatives.

\begin{table}[!htb]
    \centering
    \caption{EM and F1 scores in the test set for best published single models in the SQuAD leaderboard \cite{squad-website}}
    \begin{tabular}{| c || c | c |}
        \hline
        Model & EM (\%) & F1 (\%) \\ [0.5ex]
        \hline
        \hline
        Reinforced Mnemonic Reader \cite{Hu2017} & 79.545 &	86.654\\
        MEMEN \cite{PanLZCCH17} & 78.234 & 85.344 \\ FRC \cite{anony-nonRNN} & 76.240 & 84.599\\
        RaSoR + TR + LM  \cite{Salant2017} & 77.583 & 84.163 \\
        Stochastic Answer Networks  \cite{Xiaodong2017} & 76.828 & 84.396 \\
        r-net \cite{Yang2017} & 76.461 & 84.265 \\
        FusionNet \cite{Hsin2017} & 75.968 & 83.900 \\ DCN+ \cite{DBLP:journals/corr/abs-1711-00106}& 75.087 & 83.081\\
        Conductor-net \cite{RiuLiu2017} & 74.405 & 82.742 \\
        BiDAF + Self Attention \cite{bidaf+self-att}& 72.139 & 81.048\\
        smartnet \cite{Zhequian2017} & 71.415 & 80.160 \\
        Ruminating Reader \cite{DBLP:journals/corr/GongB17} & 70.639 & 79.456 \\
        jNet \cite{DBLP:journals/corr/ZhangZCDWJ17} &    70.607 &    79.821\\
ReasoNet \cite{DBLP:journals/corr/ShenHGC16}&	70.555 & 79.364 \\
Document Reader \cite{Chen2017} &70.733	&79.353\\
RaSoR \cite{DBLP:journals/corr/LeeKP016} &70.849 &	78.741\\
FastQAExt \cite{DBLP:journals/corr/WeissenbornWS17}&	70.849 &	78.857\\
        Multi-Perspective Matching \cite{DBLP:journals/corr/WangMHF16}&    70.387 &    78.784\\
                SEDT \cite{DBLP:journals/corr/LiuHWYN17} &    68.163&    77.527\\
       
        \textbf{FABIR (Ours)} & \textbf{67.744}
        &\textbf{77.605}\\ 
                BiDAF \cite{Seo2016} & 67.974 & 77.323 \\
                Dynamic Coattention Networks \cite{Xiong2016} & 66.233 & 77.896 \\
        Match-LSTM with Bi-Ans-Ptr \cite{Wang16} & 64.744 &	73.743\\
        Fine-Grained Gating \cite{DBLP:journals/corr/YangDYHCS16} &	62.446	& 73.327\\
        OTF dict+spelling \cite{DBLP:journals/corr/BahdanauBJGVB17} &	64.083	& 73.056\\
Dynamic Chunk Reader \cite{DBLP:journals/corr/YuZHYXZ16}&62.499&	70.956\\

        \hline
        \end{tabular}
    \label{tab:SQuAD_comparison}
\end{table}

\section{Conclusion and Future Work}
The experiments validate that attention mechanisms alone are enough to power an effective question-answering model. Above all, FABIR proved roughly five times faster at both training and inference than BiDAF, a competing RNN-based model with similar performance \cite{Seo2016}. These results strengthen some of FABIR's compelling advantages, notably, an architecture that is both more parallelizable and lighter, with half of the number of parameters in comparison to BiDAF \cite{Seo2016}.

FABIR also brings three significant contributions to this new class of neural network architectures. The convolutional attention, the reduction layer, and the column-wise cross-attention individually increased the model's F1 and EM scores by more than 2\%. Moreover, being thoroughly compatible with the Transformer \cite{Vaswani}, these new mechanisms are valuable assets to further developments in attention models. In fact, an intriguing line for future research is to evaluate their impact on other NLP tasks, such as machine translation or parsing.

Although FABIR is still far from surpassing the models at the top of the SQuAD leaderboard (Table \ref{tab:SQuAD_comparison}), we believe that its faster and lighter architecture already make it an attractive alternative to RNN-based models, especially for applications with limited processing power or that require low-latency. Also, being a distinct technique, FABIR might have low correlation with existing RNN-based models, increasing the potential of ensemble methods. How to combine FABIR with other systems is then an interesting topic for future research in diverse NLP applications.

\newpage



%
\bibliographystyle{IEEEtran}
\bibliography{IEEEabrv,Tibino.bib}

\end{document}